\documentstyle[twocolumn, art10, a4]{article}
\begin{document}

\newenvironment{eg}{\small \vspace{-0mm} \begin{flushleft} \begin{tabular}{@{\
\ \ \ \ \ }rl}}
                   {\end{tabular} \end{flushleft} \vspace{-0mm}}

\newenvironment{ex}{\small \vspace{-2.5mm} \begin{flushleft}
\begin{tabular}{rp{2.7in}}}
                   {\end{tabular} \end{flushleft} \vspace{-5mm}}

\title{\vspace{-2mm} \large EXPLORING THE ROLE OF PUNCTUATION IN PARSING
NATURAL TEXT}

\author{\normalsize  Bernard E M Jones}
\date{\normalsize Centre for Cognitive Science, University of Edinburgh,
Edinburgh EH8 9LW, Scotland\\
Email: {\sf bernie@cogsci.ed.ac.uk}}
\maketitle

\subsubsection{\hspace{4mm}ABSTRACT}
\begin{list}{}{\setlength{\leftmargin}{4mm} \setlength{\listparindent}{-4mm}
               \setlength{\parsep}{0mm} \setlength{\rightmargin}{4mm} }

\item Few, if any, current NLP systems make any significant use
of punctuation. Intuitively, a treatment of punctuation seems
necessary to the analysis and production of text. Whilst this
has been suggested in the fields of discourse structure, it is still
unclear whether punctuation can help in the syntactic field.
This investigation attempts to answer this question by parsing
some corpus-based material with two similar grammars --- one
including rules for punctuation, the other ignoring it. The
punctuated grammar significantly out-performs the
unpunctuated one, and so the conclusion is that punctuation
can play a useful role in syntactic processing.
\end{list}

\subsubsection{INTRODUCTION}

There are no current text based natural language analysis or
generation systems that make full use of punctuation, and while there
are some that make limited use, like the Editor's Assistant [Dale
1990], they tend to be the exception rather than the rule. Instead,
punctuation is usually stripped out of the text before processing, and
is not included in generated text.

Intuitively, this seems very wrong. Punctuation is such an integral
part of written language that it is difficult to imagine naturally
producing any significant body of unpunctuated text, or being able to
easily understand any such body of text.

However, this is what has been done in the computational linguistics
field. The reason that it has always been too difficult to incorporate
any coherent account of punctuation into any system is because no such
coherent account exists.

Punctuation has long been considered to be intimately related to
intonation: that is that different punctuation marks simply give the
reader cues as to the possible prosodic and pausal characteristics of
the text [Markwardt, 1942]. This claim is questioned by Nunberg
[1990], since such a transcriptional view of punctuation is
theoretically uninteresting, and also correlates rather badly with
intonation in any case.

However, even if we recognise that punctuation fulfils a linguistic
role of its own, it is by no means clear how this role is defined.
Since there is still no concise linguistic account of the function of
punctuation, we have to rely mainly on personal intuitions. This in
turn introduces new problems, since there is a great deal of
idiosyncrasy associated with the use of punctuation marks. Whilst most
people may agree on core situations in which use of a given
punctuation mark is desirable, or even necessary, there are still many
situations where their use is less clear.

In his recent review, Humphreys [1993] suggests that accounts of
punctuation fall into three categories: ``The first \ldots is
selflessly dedicated to the task of bringing Punctuation to the
Peasantry \ldots The second sort is the Style Guide, written by
editors and printers for the private pleasure of fellow professionals
\ldots The third, on the linguistics of the punctuation system, is
much the rarest of all.''

Thus whilst we do not really want to rely on publishers' style guides,
since the accounts of punctuation they contain are rather too
proscriptive and concentrate on the use of punctuation rather than its
meaning, the academic accounts of punctuation are far from numerous.
In the work of Dale [1991], the potential of punctuation in the field
of discourse and natural language generation is explored. However,
little mention is made anywhere of the role of punctuation within a
syntactic framework. Therefore the current investigation tries to
determine whether taking consideration of punctuation can further the
goals of syntactic analysis of natural language.

\subsubsection{PUNCTUATION}

Punctuation, as we consider it, can be defined as the central part of
the range of non-lexical orthography. Although arguments could be made
for including the sub-lexical marks (e.g.\ hyphens, apostrophes) and
structural marks (e.g.\ bullets in itemisations), they are excluded
since they tend to be lexicalised or rather difficult to represent,
respectively. Indeed, it is difficult to imagine the representation of
structural punctuation, other than through the use of some special
structural description language such as SGML.

Within our definition of punctuation then, we find broadly three types
of mark: delimiting, separating and disambiguating, as described by
Nunberg [1990]. Some marks, the comma especially, fall into multiple
categories since they can have different roles, and the categories
each perform distinct linguistic functions.

Delimiters (e.g.\ comma, dash, parenthesis) occur to either side of a
particular lexical expression to remove that expression from the
immediate syntactic context of the surrounding sentence (1).  The
delimited phrase acts as a modifier to the adjacent phrase instead.

\begin{eg}
(1) & John, my friend, fell over and died.
\end{eg}

Separating marks come between similar grammatical items and indicate
that the items form a list (2).  They are therefore similar to
conjunctions in their behaviour, and can sometimes replace
conjunctions in a list.

\begin{eg}
(2) & I came, I saw, I conquered.\\
    & I want butter, eggs and flour.
\end{eg}

Disambiguating marks, usually commas, occur
where an unintentional ambiguity could result if the marks were not
there (3), and so perhaps illustrate best
why the use of punctuation within NL systems could be beneficial.

\begin{eg}
(3) & Earlier, work was halted.
\end{eg}

In addition to the nature of different punctuation marks, there are
several phenomena described by Nunberg [1990] which it is useful to
consider before implementing any treatment of punctuation:
\begin{description}
\itemsep 0mm
\parsep 0mm
\parskip 0mm

\item {\it Point absorption}: strong point symbols (comma, dash,
      semicolon, etc.) absorb weaker adjacent ones (4). Commas are
      least powerful, and periods\footnote{Throughout this paper I
      shall refer to sentence-final dots as periods rather than
      full-stops, to avoid confusion.} most powerful;

\begin{eg}
(4) & It was John, my friend.
\end{eg}

\item {\it Bracket absorption}: commas and dashes are removed if they
      occur directly before an end quote or parenthesis (5);

\begin{eg}
(5) & \ldots(my brother, the teacher)\ldots
\end{eg}

\item {\it Quote transposition}: punctuation directly to the right of
      an end quote is moved to the left of that character (6). This
      phenomenon occurs chiefly in American English, but can
      occur generally;

\begin{eg}
(6) & He said, ``I love you.''
\end{eg}

\item {\it Graphic absorption}: orthographically, but not linguistically,
      similar coincident symbols are absorbed (7). Thus the dot
      marking an abbreviation will absorb an adjacent period
      whereas it would not absorb an adjacent comma.

\begin{eg}
(7) & I work for the C.I.A., not the F.B.I.
\end{eg}

\end{description}

In addition to the phenomena associated with the interaction of
punctuation, there are also distinct phenomena observable in the
interaction of punctuation and lexical expressions. Thus delimited
phrases cannot immediately contain delimited phrases of the same type
(the sole exception may be with parentheticals, though many people
object to nested parentheses) and adjuncts such as the colon-expansion
cannot contain further similar adjuncts. Therefore, in the context of
colon and semicolon scoping, (8) is ambiguous, but (9) is not.

\begin{eg}
(8) & words : words ; words .\\
(9) & words : words ; words : words .
\end{eg}

\subsubsection{THE GRAMMAR}

Recognition of punctuational phenomena does not imply that they can be
successfully encoded into a NL grammar, or whether the use of such a
punctuated grammar will result in any analytical advantages.

Nunberg [1990] advocates two separate grammars, operating at different
levels. A lexical grammar is proposed for the lexical expressions
occurring between punctuation marks, and a text grammar is proposed
for the structure of the punctuation, and the relation of those marks
to the lexical expressions they separate. The text grammar has within
it distinct levels, such as phrasal and clausal, at which distinct
punctuational phenomena can occur.

This should, in theory, make for a very neat system: the lexical
syntactic processes being kept separate from those that handle
punctuation. However, in practice, this system seems unlikely to
succeed since in order to work, the lexical expressions that occur
between punctuation marks must carry additional information about the
syntactic categories occurring at their edges so that the text grammar
can constrain the function of the punctuation marks.

For example, if a sentence includes an itemised noun phrase (10), the
lexical expression before the comma must be marked as ending with a
noun phrase, and the lexical expression after the comma must be marked
as starting with a noun phrase. A rule in the text grammar could then
process the separating comma as it clearly comes between two similar
syntactic elements.

\begin{eg}
(10) & He likes Willy, Ian and Tom.\\
     & \hspace{6mm} [end: np]\ \ [start: np]
\end{eg}

However, as (11) shows, the separating comma concept could require
information about the categories at arbitrarily deep levels occurring
at the ends of lexical expressions surrounding punctuation marks.

\begin{eg}
(11) & I like to walk, skip, and run.\\
     & I like to walk, to skip, and to run.\\
     & I like to walk, like to skip, but hate to run.
\end{eg}

Even with the above edge-category information, the parsing process is
not necessarily made any easier (since often the full partial parses
of all the separate expressions have to be held and joined). Therefore
we seem to be at no advantage if we use this approach. In addition, it
is difficult to imagine what linguistic or psychological motivation
such a separation of punctuation from lexical text could hold, since
it seems rather unlikely that people process punctuation at a separate
level to the text it surrounds.

Hence it seems more sensible to use an integrated grammar, which
handles both words and punctuation. This lets us describe the
interaction of punctuation and lexical expressions far more logically
and concisely than if the two were separated.  Good examples of this
are disambiguating commas; in a unified grammar we can simply write
rules with an optional comma among the daughters (12).

\begin{eg}
(12) & s $\rightarrow$ np (comma) vp.\\
     & s $\rightarrow$ pp (comma) s.
\end{eg}

A feature-based tag grammar was written for this investigation (based
loosely on one written by Briscoe and Waegner [1992]), and used in
conjunction with the parser included in the Alvey Tools' Grammar
Development Environment (GDE) [Carroll et al, 1991], which allows for
rapid prototyping and easy analysis of parses. It should be stressed
that this grammar is solely one of tags, and so is not very detailed
syntactically.

In order to handle the additional complications of punctuation, the
notion of {\it stoppedness} of a category has been introduced. Thus
every category in the grammar has a {\it stop} feature which describes
the punctuational character following it (13), and defaults to [st --]
(unstopped) if there is no such character.

\begin{eg}
(13) & the man,\ \ = [st c]\\
     & with the flowers.\ \ = [st f]
\end{eg}

Since the rules of the grammar further dictate that the mother
category inherits the stop value of its rightmost daughter, only rules
to specifically add punctuation for categories which could be
lexicalised are necessary. Thus a rule for the additional of a
punctuation mark after a lexicalised noun would be as in (14).  (The
calligraphic letters represent unification variables.)

\begin{eg}
(14) & n0[st $\cal S$] $\rightarrow$ n0[st --] [punc $\cal S$]
\end{eg}

We can then specify that top level categories must be [st f] (period),
that items in a list should be \mbox{[st c]} (comma), etc. In rules where we
want to force a particular punctuation mark to the right of a
category, that mark can be included in the rule, with the preceding
category unstopped: (15) illustrates the addition of a comma-delimited
noun phrase to a noun phrase. Specifically mentioning the punctuation
mark prevents the delimited phrase from being unstopped, resulting in
an unstopped mother category. Note that the phenomenon of point
absorption has been captured by unifying the value of the {\it st}\/
feature of the mother and the identity of the final punctuation mark.
Thus the possible values of {\it st}\/ are all the possible values of
{\it punc} in addition to [st --].

\begin{eg}
(15) & np[st $\cal S$] $\rightarrow$ np[st c] np[st --] [punc $\cal S$].
\end{eg}

Thus the stop feature seems sufficient to cope with the punctuational
phenomena introduced above. In order to incorporate the phenomena of
interaction between punctuation and lexical expressions (e.g.\
preventing immediate nesting of similar delimited phrases), we need to
introduce a small number of additional features into the grammar. If,
for example, we make a comma-delimited noun phrase [cm +], we can then
stipulate that any noun phrase that includes a comma-delimited phrase
has the feature [cm --], so that the two cannot unify (16). Note that
the unification of mother and right-most daughter stop values is
omitted for clarity of presentation.

\begin{eg}
(16) & np[cm --] $\rightarrow$ np[st c] np[cm +, st --]
\end{eg}

We can incorporate the relative scoping of colons and semicolons, as
discussed previously, into the grammar very easily too. The semicolon
rule (17) accepts any value of {\it co}\/ in its arguments, but the
colon rule (18) only accepts [co --]. The mother category of the colon
rule bears the feature [co +] to prevent inclusion into further
colon-bearing sentences. Note that there are more versions of the
colon rule, which deal with different constituents to either side of
the colon, and also that, since the GDE does not permit the
disjunction of feature values, the semicolon rule is merely an
abbreviation of the multiple rules required in the grammar. Stop
unification is again omitted.


\begin{eg}
(17) & s[co ($\cal A \vee \cal B$)] $\rightarrow$
     s[co $\cal A$, st sc] s[co $\cal B$].\\
(18) & s[co +] $\rightarrow$ s[co --, st co] s[co --].
\end{eg}


Hence the inclusion of a few simple extra features in a normal grammar
has achieved an acceptable treatment of punctuational phenomena. Since
this work only represents the initial steps of providing a full and
proper account of the role of punctuation, no claims are made for the
theoretical validity or completeness of this approach!

\subsubsection{THE CORPUS}

For the current investigation it was necessary to use a corpus
sufficiently rich in punctuation to illustrate the possible advantages
or disadvantages of utilising punctuation within the parsing process.
Obviously a sentence which includes no punctuation will be equally
difficult to parse with both punctuated and unpunctuated grammars.
Similarly, for sentences including only one or two marks of
punctuation, the use of punctuation is likely to be rather procedural,
and hence not necessarily very revealing.

Therefore the tagged Spoken English Corpus was chosen [Taylor \&
Knowles, 1988]. This features some very long sentences, and includes
rich and varied punctuation. Since the corpus has been punctuated
manually, by several different people, some idiosyncrasy occurs in the
punctuational style, but there is little punctuation which would be
deemed inappropriate to the position it occurs in.

A subset of 50 sentences was chosen from the whole corpus.  Between
them these sentences include material taken from news broadcasts,
poetry readings, weather forecasts and programme reviews, so a wide
variety of language is covered.

The lengths of the sentences varied from 3 words to 63 words, the
average being 31 words; and the punctuational complexity of the
sentences varied from one mark (just a period) to 16 marks, the
average being 4 punctuation marks.  A sample tagged sentence is shown
in (19), where {\it fs}\/ denotes a period.

\begin{ex}
(19) & {\centering Their\_APP\$ meeting\_NN1 involves\_VVZ a\_AT1
        kind\_NN1 of\_IO life\_NN1 swap\_NN1 fs\_FS \\}
\end{ex}

The punctuated grammar, developed with this subset of the corpus, was
used to parse the corpus subset, and then an unpunctuated version of
the same grammar was used to parse the same subset. The reason that
testing was performed on the training corpus was that, in the absence
of a complete treatment of punctuation, the punctuational phenomena in
the training corpus were the only ones the grammar could work with,
and although they included almost all of the core phenomena mentioned,
slightly different instances of the same phenomena could cause a parse
failure. For reference, a small set of novel sentences were also
parsed with the grammars, to determine their coverage outside the
closed test.

The unpunctuated version of the grammar was prepared by removing all
the features relating to specifically punctuational phenomena, and
also removing explicit mention of punctuation marks from the rules.
This, of course, left behind certain rules that were functionally
identical, and so duplicate rules were removed from the grammar.
Similarly for rules which performed the same function at different
levels in the grammar (e.g.\ attachment of prepositions to the end of
a sentence with a comma was also catered for by rules allowing
prepositions to be attached to noun and verb phrases without a
comma).

\subsubsection{RESULTS}

Results of parsing with the punctuated grammar were very good,
yielding, on average, a surprisingly small number of parses. The
number of parses ranged from 1 to 520, with an average of 38. This
average is unrepresentatively high, however, since only 4 sentences
had over 50 parses. These were, in general, those with high numbers of
punctuation marks, all containing at least 5, as in (20). Ignoring the
four smallest and four largest results then, the average number of
parses is reduced to just 15. Example (21) is more representative of
parsing. On examination, a great number of the ambiguities seem to be
due to inaccuracies or over-generality in the lexical tags assigned to
words in the corpus. The word {\it more}, for example, is triple
ambiguous as determiner, adjective and noun, irrespective of where it
occurs in a sentence.

\begin{ex}
(20) & {\centering (The sunlit weeks between were full of maids:
        Sarah, with orange wig and horsy teeth, was so bad-tempered
        that she scarcely spoke; Maud was my hateful nurse who smelled
        of soap, and forced me to eat chewy bits of fish, thrusting me
        back to babyhood with threats of nappies, dummies, and the
        feeding bottle.)\\ 520 punctuated parses\\}
\vspace{-2mm}\\
(21) & {\centering (More news about the reverend Sun Myung Moon, founder
        of the Unification Church, who's currently in jail for tax
        evasion: he was awarded an honorary degree last week by the
        Roman Catholic University of la Plata in Buenos Aires,
        Argentina.) 18 punctuated parses\\}
\end{ex}

Besides the ambiguity of corpus tags, a problem arose with
words that had been completely mistagged. If these caused the
parse to fail completely, the tag was changed in the development
phase of the grammar, but even so, the number of complete
mistags was rather small in the sub-corpus used: around 10
words in the 50 sentences used.

Initial attempts at parsing the corpus subset using the unpunctuated
version of the grammar were unsuccessful on even the most powerful
machine available. This was due to the failure of the machine to
represent all the parses simultaneously when unpacking the parse
forest produced by the chart parser. A special section of code written
for the GDE (grateful thanks are due to John Carroll for supplying
this piece of code) to estimate the number of individual parses
represented by the packed parse-forest showed that for all but the
most basically punctuated sentences, the number of parses was
ridiculously huge. The figure for the sentence in (21) was in excess
of 6.3$\times$10$^{\tiny 12}$ parses! Even though this estimate is an
upper bound, since effects of feature value percolation during
unpacking are ignored, it has been fairly accurate with most grammars
in the past and still indicates that rather too many parses are being
produced!  Not all sentences produced such a massive number of parses:
the sentence in (22) yielded only 192 parses with the unpunctuated
grammar which was by far the smallest number of unpunctuated parses.
Most sentences that managed to pass the estimation process produced
between 10$^{\tiny 6}$ and 10$^{\tiny 9}$ parses.

\begin{ex}
(22) & {\centering (Protestants, however, are a tiny minority in
        Argentina, and the delegation won't be including a Roman
        Catholic.)\\ 9 punctuated parses \\}
\end{ex}

On examination of the grammar and the corpus, it is possible to
understand why this has happened. The punctuated grammar had to allow
for sentences including comma-delimited noun phrases adjacent to
undelimited noun phrases, as illustrated by the rules (15) and (16).
These are relatively easy to mark and recognise when the punctuation
is available. However, without punctuational clues, and with the
under-specific tagging system, any compound noun could appear as a set
of delimited noun phrases with the unpunctuated grammar.

Therefore the unpunctuated grammar was further trimmed, to such an
extent that parses no longer accurately reflected the linguistic
structure of the sentences, since, for example, comma delimited noun
phrases and compound nouns became indistinguishable. Some manual
preparation of the sentences was also carried out to prevent the
reoccurrance of simple, but costly, misparses.

The results of the parse now became much more tractable. For basic
sentences, as predicted, there was little difference in the
performance of punctuated and unpunctuated grammars.  Results were
within an order of magnitude, showing that no significant advantage was
gained through the use of punctuation. The sentences in (23) and (24)
received 1 and 11 parses respectively with the unpunctuated grammar.

\begin{ex}
(23) & {\centering (Well, just recently, a day conference on miracles
        was convened by the research scientists, Christian Fellowship.)\\
        4 punctuated parses \\}
\vspace{-2mm}\\
(24) & {\centering (The assembly will also be discussing the UK
        immigration laws, Hong Kong, teenagers in the church, and
        of course, church unity schemes.)\\ 2 punctuated parses\\}
\vspace{-2mm}\\
(25) & {\centering (They want to know whether, for instance, in a
        scientific age, Christians can really believe in the story
        of the feeding of the five thousand as described, or was the
        miracle that those in the crowd with food shared it with those
        who had none?) 24 punctuated parses\\}
\end{ex}

For the most complex sentences, however, the number of parses with the
unpunctuated grammar was typically more than two orders of magnitude
\mbox{higher} than with the punctuated grammar. The sentence in (25) had
12,096 unpunctuated parses.

Parsing a set of ten previously unseen punctuationally complex
sentences with the punctuated grammar resulted in seven of the ten
being unparsable. The other three parsed successfully, with the number
of parses falling within the range of the results of the first part of
the investigation. The parse failures, on examination, were due to
novel punctuational constructions occurring in the sentences which the
grammar had not been designed to handle.  Parsing the unseen sentences
with the unpunctuated grammar resulted in one parse failure, with the
results for the other 9 sentences reflecting the previous results for
complex sentences.

\subsubsection{DISCUSSION}

This investigation seems to support the original premise --- that
inclusion and use of punctuational phenomena within natural language
syntax can assist the general aims of natural language processing.

We have seen that for the simplest sentences, use of punctuation gives
us little or no advantage over the more simple grammar, but,
conversely, does no harm and can reflect the actual linguistic
construction a little more accurately.

For the longer sentences of real language, however, a grammar which
makes use of punctuation massively outperforms an otherwise similar
grammar that ignores it. Indeed, it is difficult to see how any
grammar that takes no notice of punctuation could ever become
successful at analysing such sentences unless some huge amount of
semantic and pragmatic knowledge is used to disambiguate the analysis.

However, as was shown by the attempt at parsing the novel sentences,
knowledge of the role of punctuation is still severely limited. The
grammar only performed reliably on those punctuational phenomena it
had been designed with.  Unexpected constructs \mbox{caused} it to
fail totally. Therefore, following the recognition that punctuation
can play a crucial role in natural language syntax, what is needed is
a thorough investigation into the theory of punctuation. Then
theoretically based analyses of punctuation can play a full and
important part in the analysis of language.

\subsubsection{ACKNOWLEDGEMENTS}

This work was carried out under Esprit \mbox{Acquilex-II}, BRA 7315,
and an ESRC Research Studentship, R00429334171.  Thanks for
instructive and helpful comments to Ted Briscoe, John Carroll, Robert
Dale, Henry Thompson and anonymous CoLing reviewers.

\subsubsection{REFERENCES}

\begin{list}{}{\setlength{\leftmargin}{5mm} \setlength{\listparindent}{-5mm}
               \setlength{\parsep}{0mm}}

\item \hspace{-5mm}Briscoe, E J and N Waegner (1992). ``Robust Stochastic
Parsing Using the Inside-Outside Algorithm.'' In {\it Proceedings, AAAI
Workshop on Statistically-based NLP Techniques}, San Jose, CA.

Carroll, J; E J Briscoe; and C Grover (1991). ``A Development
Environment for Large Natural Language Grammars.'' Technical
Report 233, Cambridge University Computer Laboratory.

Dale, R (1991). ``Exploring the Role of Punctuation in the Signalling
of Discourse Structure.'' In {\it Proceedings, Workshop on Text
Representation and Domain Modelling}, T.\ U.\ Berlin, pp110-120.

Dale, R (1990). ``A Rule-based approach to Computer-Assisted Copy
Editing.'' {\it Computer Assisted Language Learning}, {\bf 2}, pp59--67.

Humphreys, R L (1993). ``Book Review: The Linguistics of
Punctuation.'' {\it Machine Translation}, {\bf 7}.

Markwardt, A H (1942). {\it Introduction to the English Language},
Oxford University Press, New York.

Nunberg, G (1990). {\it The Linguistics of Punctuation}, CSLI Lecture
Notes {\bf 18}, Stanford, CA.

Taylor, L J and G Knowles (1988). {\it Manual of Information to
Accompany the SEC Corpus}, University of Lancaster.
\end{list}

\end{document}